\documentclass[11pt,a4paper]{article}
\usepackage[hyperref]{eacl2021}
\usepackage{times}
\usepackage{latexsym}

\usepackage{booktabs}
\usepackage{graphicx}
\usepackage{amsmath}
\usepackage{amssymb}
\DeclareMathOperator*{\argmax}{arg\,max}
\usepackage{bm}
\usepackage[ruled,vlined,linesnumbered]{algorithm2e}
\usepackage{hyperref}
\usepackage{multirow}

\usepackage{microtype}

\aclfinalcopy % Uncomment this line for the final submission
\title{Removing Word-Level Spurious Alignment between Images and Pseudo-Captions in Unsupervised Image Captioning}

\author{Ukyo Honda$^{1,3}$\qquad Yoshitaka Ushiku$^2$\qquad Atsushi Hashimoto$^2$\\
\textbf{Taro Watanabe}$^1$\qquad
\textbf{Yuji Matsumoto}$^3$\\
$^1$ Nara Institute of Science and Technology\qquad
$^2$ OMRON SINIC X Corp.\\ $^3$ RIKEN Center for Advanced Intelligence Project\\
$^1$ {\texttt{\{honda.ukyo.hn6, taro\}@is.naist.jp}}\\
$^2$ {\texttt{\{yoshitaka.ushiku, atsushi.hashimoto\}@sinicx.com}}\\ $^3$ \texttt{yuji.matsumoto@riken.jp}
}

\date{}

\begin{document}
\maketitle
\begin{abstract}
Unsupervised image captioning is a challenging task that aims at generating captions without the supervision of image--sentence pairs, but only with images and sentences drawn from different sources and object labels detected from the images.
In previous work, \textit{pseudo-captions}, {\em i.e.}, sentences that contain the detected object labels, were assigned to a given image.
The focus of the previous work was on the alignment of input images and pseudo-captions at the sentence level.
However, pseudo-captions contain many words that are irrelevant to a given image.
In this work, we investigate the effect of removing mismatched words from image--sentence alignment to determine how they make this task difficult.
We propose a simple gating mechanism that is trained to align image features with only the most reliable words in pseudo-captions: the detected object labels.
The experimental results show that our proposed method outperforms the previous methods without introducing complex sentence-level learning objectives.
Combined with the sentence-level alignment method of previous work, our method further improves its performance.
These results confirm the importance of careful alignment in word-level details.\footnote{Code will be available at \url{https://github.com/ukyh/RemovingSpuriousAlignment}}
\end{abstract}

\begin{figure*}[t]
    \centering
    \includegraphics[width=1.0\textwidth,keepaspectratio]{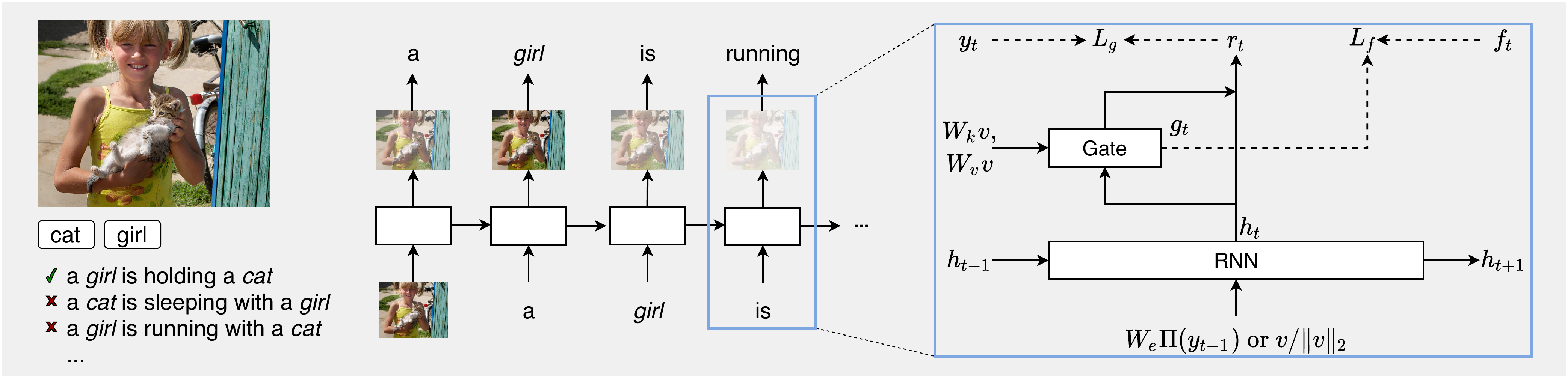}
    \caption{Overview of our model. The input is listed on the left-hand side: an image, its detected object labels, and its pseudo-captions. The model learns to generate the pseudo-captions while considering the correspondence between the image and each word being generated. The detailed process is shown in the blue box on the right-hand side. The base encoder--decoder model output $\bm{h}_t$, a gate value $g_t$, and a pseudo-label $f_t$ on the gate are described in Sections~\ref{encdec}, \ref{gate}, and \ref{filter}, respectively. The dashed arrows indicate the processes conducted only during training.}
    \label{fig:gate}
\end{figure*}

\section{Introduction}

Image captioning is a task to describe images in natural languages.
This is a fundamental challenge with regard to automatically retrieving and summarizing the visual information in a human-readable form.
Recently, considerable progress has been made \cite{vinyals2015, xu2015, anderson2018} owing to the development of neural networks and a large number of annotated image--sentence pairs \cite{young2014, lin2014, krishna2017}.
However, these pairs are limited in their coverage of scenes\footnote{For example, the standard captioning dataset MS COCO \cite{lin2014} covers only approximately 100 object categories out of 500 object categories defined in an object detection dataset \cite{agrawal2019nocaps}. 
In addition to objects, attributes and relations are also not covered well owing to the small vocabulary size, 8791 \cite{karpathy2015}.}, and scaling them is difficult owing to the cost of manual annotation.

Unsupervised image captioning \cite{feng2019} aims to describe scenes that have no corresponding image--sentence pairs, without requiring additional annotation of the pairs.
The only available resources are images and sentences drawn from different sources and object labels detected from the images.
Although it is highly challenging, unsupervised image captioning has the potential to cover a broad range of scenes by exploiting a large number of images and sentences that are not paired by expensive manual annotation.

To train a captioning model in this setting, previous work \cite{feng2019, laina2019} employed sentences that contained the object labels detected from given images.
We refer to these sentences as \textit{pseudo-captions}.
However, pseudo-captions are problematic in that they are likely to contain words that are irrelevant to the given images.
Assume that an image contains two objects \textit{cat} and \textit{girl} (Figure~\ref{fig:gate}).
This situation could give rise to various possible pseudo-captions, {\em e.g.}, ``a girl is holding a cat,'' ``a cat is sleeping with a girl,'' ``a girl is running with a cat.''
When the first sentence is the correct caption of the image, the words \textit{sleeping} and \textit{running} of the other sentences are irrelevant to the image.
As the detected object labels provide insufficient information to judge which sentence corresponds to the image, many pseudo-captions containing such mismatched words can be produced.

Regardless of the problem in pseudo-captions, previous work \cite{feng2019,laina2019} did not explicitly remove word-level mismatches.
They tried aligning the features of images and their pseudo-captions at the sentence level.
Although this line of approach can potentially align the images and sentences correctly if there are sentences that exactly describe each image, it is not likely to hold for the images and sentences retrieved from different sources.

To shed light on the problem of word-level spurious alignment in the previous work, we focus on removing mismatched words from image--sentence alignment.
To this end, we introduce a simple gating mechanism that is trained to exclude image features when generating words other than the most reliable words in pseudo-captions: the detected objects.
The experimental results show that the proposed method outperforms previous methods without introducing complex sentence-level learning objectives.
Combined with the sentence-level alignment method of previous work, our method further improves its performance.
These results confirm the importance of careful alignment in word-level details.

\section{Method}

Our model comprises a sequential encoder--decoder model, a gating mechanism on the encoder--decoder model, a pseudo-label on the gating mechanism, and a decoding rule to avoid the repetition of object labels, as presented in Figure~\ref{fig:gate}.

\subsection{Base Encoder--Decoder Model}
\label{encdec}

Typical supervised, encoder--decoder captioning models maximize the following objective function during training:
\begin{equation}
    \label{eq:sup encoder--decoder}
    \bm{\theta}^* = \argmax_{\bm{\theta}} \sum_{(\bm{I}, \bm{y})} \log p(\bm{y} | \bm{I}; \bm{\theta}),
\end{equation}
where $\bm{\theta}$ are the parameters of the models, $\bm{I}$ is a given image, and $\bm{y} = y_1, ..., y_T$ is its corresponding caption, the last token $y_T$ is a special end-of-sentence token.

However, in unsupervised image captioning, the corresponding caption $\bm{y}$ is not available.
Instead, object labels in given images are provided by pre-trained object detectors.
Previous work utilized the detected object labels to assign a roughly corresponding caption $\bm{\hat{y}}$, {\em i.e.}, a pseudo-caption, to the given image.
Following the previous work, we define pseudo-captions of an image as sentences containing the object labels detected from the image.
Given the pseudo-caption $\bm{\hat{y}}$, our base encoder--decoder model maximizes the following objective function:
\begin{equation}
    \label{eq:unsup encoder--decoder}
    \bm{\theta}^* = \argmax_{\bm{\theta}} \sum_{(\bm{I}, \bm{\hat{y}})} \log p(\bm{\hat{y}} | \bm{I}; \bm{\theta}).
\end{equation}

In encoder--decoder captioning models, the probability $p(\bm{y} | \bm{I})$\footnote{Hereafter, we omit the model parameter $\bm{\theta}$ for brevity.} is auto-regressively factorized as $p(\bm{y} | \bm{I}) = \prod_{t = 1}^T p(y_t | y_{<t}, \bm{I})$ and each $p(y_t | y_{<t}, \bm{I})$ is computed by a single step of recurrent neural networks (RNNs).
The encoder encodes the given image $\bm{I}$ to an image representation $\bm{v} \in \mathbb{R}^{d'}$ that is fed to the decoder as an initial input to generate a sequence of words auto-regressively.
The detailed computation of $p(\hat{y}_t | \hat{y}_{<t}, \bm{I})$ is as follows:
\begin{align}
    \label{eq:base}
    p(\hat{y}_t | \hat{y}_{<t}, \bm{I}) &= \frac{\textrm{exp}(\bm{h}_{t}^\top \bm{W}_o \Pi (\hat{y}_{t}))}{\sum_{y' \in \mathcal{Y}} \textrm{exp}(\bm{h}_t^\top \bm{W}_o \Pi (y'))}, \\
    \bm{h}_t &= 
    \begin{cases}
        \textrm{Dec} \left( \frac{\bm{v}}{\| \bm{v} \|_2}, \bm{h}_0 \right), & \text{if } t = 1; \\
        \textrm{Dec}(\bm{e}_t, \bm{h}_{t - 1}), & \text{otherwise},
    \end{cases} \\
    \bm{v} &= \bm{W}_a \textrm{Enc}(\bm{I}), \\
    \bm{e}_t &= \bm{W}_e \Pi (\hat{y}_{t - 1}),
\end{align}
where $\textrm{Enc}(\cdot)$ is a pre-trained image encoder with a linear transformation matrix $\bm{W}_a \in \mathbb{R}^{d \times d'}$ on top of it, $\textrm{Dec}(\cdot)$ is an RNN decoder, $\Pi(\cdot)$ is the one-hot encoding function, $\bm{h}_0 \in \mathbb{R}^{d}$ is a zero vector, $\mathcal{Y}$ is the whole vocabulary to use, and $\bm{W}_e, \bm{W}_o \in \mathbb{R}^{d \times |\mathcal{Y}|}$ are the word embedding matrices.
Details of the encoder and decoder are provided in Section~\ref{implement}.

\subsection{Gating Mechanism to Consider Word-Level Correspondence}
\label{gate}

As indicated in Eq.~\ref{eq:unsup encoder--decoder}, our base encoder--decoder model decodes all of the words in pseudo-captions from the images.
However, pseudo-captions are highly likely to contain words that are irrelevant to the given images.
Thus, forcing a model to decode the pseudo-captions in their entirety from the images might be more disadvantageous than beneficial for training precise captioning models.

To enable our model to handle word-level mismatches, we introduce a simple gating mechanism.
Our model, which is equipped with this gating mechanism, takes an image representation at each $t$-th time step.
The gate is designed to control the amount of image representation used to generate the $t$-th word.
In other words, the gate is expected to determine the extent to which the given image corresponds to the $t$-th word.
With a slight modification to Eq.~\ref{eq:base}, our model with the gating mechanism is defined as follows:
\begin{align}
    \label{eq:gated prob}
    p(\hat{y}_t | \hat{y}_{<t}, \bm{I}) &= \frac{\textrm{exp}(\bm{r}_{t}^\top \bm{W}_o \Pi (\hat{y}_{t}))}{\sum_{y' \in \mathcal{Y}} \textrm{exp}(\bm{r}_t^\top \bm{W}_o \Pi (y'))}, \\
    \label{eq:gated value}
    \bm{r}_{t} &= g_t \frac{\bm{W}_v \bm{v}}{\| \bm{W}_v \bm{v} \|_2} + (1 - g_t) \bm{h}_{t}, \\
    g_t &= \textrm{sigmoid}(\textrm{tanh} (\bm{W}_{k} \bm{v})^{\top} \bm{h}_t),
\end{align}
where $\bm{W}_k,\bm{W}_v \in \mathbb{R}^{d \times d}$ are the linear transformation matrices for computing the gate value $g_t \in [0, 1]$ and the output of the gate $\bm{r}_t \in \mathbb{R}^{d}$.
When $g_t$ is close to one, it forces the model to use more information from the image ($\bm{v}$) to generate the $t$-th word; when $g_t$ is close to zero, it forces the model to do the opposite.

The fed image representation $\bm{W}_v \bm{v}$ is kept constant at every time step $t$.
Thus, even when the $t$-th word is correctly pictured in the image $\bm{I}$,  $\bm{W}_v \bm{v}$ itself cannot determine which specific object in the image should be generated according to the current context in the output caption.
Therefore, we apply L2 normalization to the image representation in Eq.~\ref{eq:gated value} to ensure that a relatively greater amount of the contextual information ($\bm{h}_t$) is used.

To train our model with the gating mechanism, we minimize the following cross-entropy loss for each pair of images and their pseudo-captions:
\begin{equation}
    \label{eq:gate loss}
        \mathcal{L}_{g} = - \frac{1}{T} \sum_{t = 1}^T \log p(\hat{y}_t | \hat{y}_{<t}, \bm{I}).
\end{equation}

\begin{table*}[t]
\centering
\scalebox{0.745}{
\begin{tabular}{lcccc}  
\toprule
 & Training Text & Object Detector & Image Encoder & Text Decoder \\
\midrule
A \cite{feng2019} & SS & Faster-RCNN trained on OpneImages-v2 & Inception-v4 & 1-layer LSTM of 512 dimensions \\
B \cite{laina2019} & GCC & Faster-RCNN trained on OpneImages-v4 & ResNet-101 & 1-layer GRU of 200 dimensions \\
\bottomrule
\end{tabular}
}
\caption{Summary of the difference in the experimental settings.}
\label{tab:settings}
\end{table*}

\subsection{Pseudo-Labels on Gate to Remove Word-Level Spurious Alignment}
\label{filter}

The above gate is expected to reflect the correspondence between images and words in pseudo-captions.
However, learning to reflect the correspondence correctly is difficult for the gate under the noisy and weak supervision of pseudo-captions.

In this work, our focus is to remove the spurious alignment between images and words in pseudo-captions.
Consequently, we apply the following rule to the gate that largely suppresses image representations to use: $g_t$ should be close to one if the $t$-th word to generate is a detected object label; otherwise, it should be close to zero.
This is based on the assumption that, given an image and its pseudo-caption, the reliable words in the pseudo-caption are only the detected object labels, and the others are likely to be irrelevant to the image.

We assign a pseudo-label $f \in \{ 0, 1 \}$ on the gate: $f_t = 1$ if the word $\hat{y}_t$ corresponds to any of the object labels detected from a given image; otherwise, $f_t = 0$.
The gate then learns the correspondence by minimizing the following loss function:
\begin{equation}
    \label{eq:filter}
        \mathcal{L}_{f} = - \frac{1}{T} \sum_{t = 1}^T \Big[ \alpha f_t \log g_t + (1 - f_t) \log (1 - g_t) \Big],
\end{equation}
where $\alpha$ is the weight to emphasize the loss when $f_t = 1$.
A relatively large value is recommended for $\alpha$ to prevent $g_t$ from always being zero because the number of detected object labels (where $f_t = 1$) in pseudo-captions is generally smaller than the number of the other words (where $f_t = 0$).

Combined with the loss function of Eq.~\ref{eq:gate loss}, the final loss function is defined as follows:
\begin{equation}
    \label{eq:total loss}
        \mathcal{L} = \mathcal{L}_{g} + \mathcal{L}_{f}.
\end{equation}

\subsection{Unique-Object Decoding}
\label{unique}

An evaluation of our model revealed that it tends to repeat words in object categories.
Although repetition is common in encoder--decoder models, this repetition was generated owing to a different cause.
As mentioned in Section~\ref{gate}, the image representation $\bm{v}$ cannot correctly predict the word $\hat{y}_t$ without the context representation $\bm{h}_t$; if the gate value $g_t$ is exactly one, the model always outputs the most salient object label in the given image.

To avoid ignoring contextual information, we applied a simple decoding rule during the evaluation.
Given that the model generates a word $ y_t$ at $t$-th time step, our decoding rule checks whether $y_t$ is in predefined object categories, {\em i.e.}, object categories defined for object detectors.
If $y_t$ is found in the object categories, the rule forces the probability of generating $y_t$ to be zero in the subsequent time steps.

% \begin{algorithm}[t]
% \caption{Unique-Object Decoding}
% \label{alg:decoding}
% \SetAlgoLined
% \KwIn{An image ${\bm I}$, trained captioning model ${\bm M}$, maximum decoding length $T$, a set of indexed object categories ${\mathcal O}$ and the index $e$ of the end-of-sentence token}
%     $decodedCaption \leftarrow$ empty list $[]$ \\
%     $decodedObjects \leftarrow \emptyset$ \\
%     % Initialize: Create an empty array $decodedCaption$ and an empty set $decodedObjects$ \\
%     $y_0 = {\bm I}$ \\
%     \For{$t = 1, \dots, T$}{
%         ${\bm p}_t = {\bm M} (y_{<t}, {\bm I})$ \\
%         \For{$n = 1, \dots, N$}{ \label{lst:line:remove begin}
%             \If{$n \in decodedObjects$}{
%                 ${\bm p}_{t, n} \leftarrow 0$ \label{lst:line:remove}
%             }
%         } \label{lst:line:remove end}
%         $y_t = \argmax_{n \in \{1, \dots, N\}} {\bm p}_t$ \\ \label{lst:line:y_t}
%         % $decodedCaption.append(y_{t})$ \\
%         $decodedCaption \leftarrow decodedCaption + [y_{t}]$ \\
%         \If{$y_t == e$}{
%             break
%         }
%         \If{$y_t \in {\mathcal O}$}{ \label{lst:line:check begin}
%             % $decodedObjects.add(y_{t})$
%             $decodedObjects \leftarrow decodedObjects \cup y_{t}$
%         } \label{lst:line:check end}
%     }
%     \KwRet{$decodedCaption$}
% \end{algorithm}

\section{Experiments}

We ran the experiments under two different settings, \newcite{feng2019} and \newcite{laina2019}, for a fair comparison with each.
For brevity, we refer to the settings in \newcite{feng2019} and \newcite{laina2019} as setting A and B, respectively.
The difference of the settings is clarified in Table~\ref{tab:settings}.

\begin{table*}[t]
\centering
\scalebox{0.78}{
\begin{tabular}{clcccccccc}  
\toprule
\multicolumn{1}{c}{} & & BLEU-1 & BLEU-2 & BLEU-3 & BLEU-4 & METEOR & ROUGE-L & CIDEr & SPICE \\
\midrule
\multirow{2}{*}{A} & \newcite{feng2019} & 41.0 & 22.5 & 11.2 & 5.6 & 12.4 & 28.7 & 28.6 & 8.1 \\
% \cline{2-10}
                   & Ours & \textbf{49.5 $\pm$ 0.7} & \textbf{27.3 $\pm$ 1.2} & \textbf{13.1 $\pm$ 0.8} & \textbf{6.3 $\pm$ 0.5} & \textbf{14.0 $\pm$ 0.1} & \textbf{34.5 $\pm$ 0.3} & \textbf{31.9 $\pm$ 1.0} & \textbf{8.6 $\pm$ 0.2} \\
% \midrule
\midrule\midrule
\multirow{2}{*}{B} & \newcite{laina2019} &   &   &   & 6.5 & 12.9 & 35.1 & 22.7 &  \\
% \cline{2-10}
                   & Ours & 50.4 $\pm$ 1.5 & 29.5 $\pm$ 0.8 & 14.4 $\pm$ 0.5 & \textbf{7.6 $\pm$ 0.4} & \textbf{13.5 $\pm$ 0.3} & \textbf{37.3 $\pm$ 0.2} & \textbf{31.8 $\pm$ 0.7} & 8.4 $\pm$ 0.1 \\
\bottomrule
\end{tabular}
}
\caption{Comparison with the state-of-the-art results on the experimental settings A and B. The scores of our model are the \textit{mean} $\pm$ \textit{standard deviation} of five runs. The scores obtained for BLEU-1 to 3 and SPICE are not provided in the original paper of \newcite{laina2019}.}
\label{tab:sota comparison}
\end{table*}

\subsection{Datasets}

\noindent \textbf{Evaluation Set.}
To evaluate our proposed method, we used the MS COCO dataset \cite{lin2014} with the validation/test split defined by \newcite{karpathy2015}.
Each split has 5,000 images and five reference captions for each image.

\noindent \textbf{Training Images.}
We used the images (without their captions) in the remaining training split of MS COCO (113,286 images), and a pre-traind object detector \cite{huang2017} to retrieve the object labels found in the images\footnote{Although this pre-trained object detector requires bounding box and semantic label annotations, it can be replaced with any multi-label image classifier, which can be trained on image-tag pairs that are largely and freely available on the web. To ensure this compatibility, bounding box features are not used in unsupervised image captioning.}.
The detector is a publicly available Faster-RCNN model\footnote{\url{https://github.com/tensorflow/models/tree/master/research/object_detection}} \cite{ren2015}.
The training data of the object detector differs depending on the previous work; thus, we used the object detector trained on OpenImages-v2 \cite{krasin2017} to compare with \newcite{feng2019} and that trained on OpenImages-v4 \cite{kuznetsova2020} to compare with \newcite{laina2019}.
Note that these object detectors were not trained on MS COCO images.
Following the previous work, we refrained from using the detected bounding boxes and their features.

\noindent \textbf{Training Text.}
Following the previous work, we used the Shutterstock image description corpus (SS) \cite{feng2019} and the training split captions (without images) of Google's Conceptual Captions (GCC) \cite{sharma2018} for comparison with \newcite{feng2019} and \newcite{laina2019}, respectively.
SS consists of 2.3M image descriptions crawled from Shutterstock, an online stock photography website; GCC consists of 3.3M image descriptions crawled from the web.
Note that these sentences are not the descriptions of the images in MS COCO.

\subsection{Implementation Details}
\label{implement}
\noindent \textbf{Image Encoder.}
For a fair comparison with the previous work, we employed different image encoders depending on the compared method: Inception-v4 \cite{szegedy2017} in the settings of \newcite{feng2019} and ResNet-101 \cite{he2016res,he2016res2} in the settings of \newcite{laina2019}.
Both image encoders were pre-trained on ImageNet \cite{russakovsky2015} and are publicly available\footnote{\url{https://github.com/tensorflow/models/tree/master/research/slim}}. 
The parameters of the image encoder were fixed during training and prediction.

\noindent \textbf{Text Decoder.}
Similar to the image encoder, we used a different RNN as our decoder: LSTM \cite{hochreiter1997} and GRU \cite{cho2014} to enable us to compare our results with those of \newcite{feng2019} and \newcite{laina2019}, respectively.
Following the previous work, the number of hidden layers' dimensions was set to 512 for LSTM and 200 for GRU.
The number of the RNN layer was set to one.
Word embeddings were randomly initialized and had the same dimensions as the RNN hidden layer.

\noindent \textbf{Pseudo-Captions.}
Captions tend to describe salient objects, not all detected objects.
For example, the frequent object \textit{person} often co-occurs with \textit{face} and \textit{clothing} in images, but these three are not always the salient objects to be described in a caption.
To avoid collecting the pseudo-captions that only contain these frequent objects, we picked up each detected object and their pairs to retrieve pseudo-captions, rather than using all detected objects.
In this retrieval, we converted object labels to their plural forms using a dictionary used in \newcite{feng2019} so that the pseudo-captions could also cover the plural forms of the objects.

\noindent \textbf{Pseudo-Caption Preprocessing.}
For each pair of objects, we selected sentences where $1 < n \leq 4$ words existed between the objects ($n$ is the number of words).
$n > 1$ is to collect neither the objects' compound words nor the sentences omitted articles, {\em e.g.}, ``\textit{plant} on \textit{table}''; $n \leq 4$ is to pick up the sentences likely to describe the relations of the target objects.
For each object, we selected sentences wherein $n \leq 2$ words were in between the object and its dependent adjective to pick up the sentences likely to describe the object in detail.
We used spaCy\footnote{\url{https://spacy.io}} en\_core\_web\_lg model for parsing.

\noindent \textbf{Value of $\alpha$.}
As described above, each pseudo-caption contains only one or two detected objects, which is very few compared with the average sentence lengths of the text corpora (12.0 in SS and 10.7 in GCC).
To balance the label imbalance of $f_t$, we searched the value for $\alpha$ (Eq.~\ref{eq:filter}) at a power of 2 and found that $\alpha = 16$, which roughly equals the quotient of $\frac{\textrm{Sentence\ Length}}{\textrm{Detected\ Objects}}$, worked well across the settings.

\noindent \textbf{Training Iteration.}
After collecting the pseudo-captions, we created a set of the objects and pairs that were used to collect the pseudo-captions.
The training is iterated over the pairs in this set, rather than over each image, to avoid overfitting for the most frequent object labels.
On each iteration of the pairs of objects, we randomly sampled the image and pseudo-caption, wherein both of the objects were contained.
Likewise, we did the same sampling on each object in the pairs.
The number of the object pairs was 11,607 and 10,612 in the settings A and B, respectively.
We set the batch size to eight and terminated the training when the best validation score (specifically, the CIDEr score) did not exceed for 20 epochs.
For the optimizer, we used Adam with the recommended hyperparameters \cite{kingma2015}.

\noindent \textbf{Evaluation.}
In the evaluation, we set the maximum decoding length to 20.
Our model decoded captions by using greedy search and unique-object decoding, described in Section~\ref{unique}.
The evaluation metrics we used were BLEU \cite{papineni2002}, ROUGE \cite{lin2004}, METEOR \cite{denkowski2014}, CIDEr \cite{vedantam2015} and SPICE \cite{anderson2016}.

\begin{table*}
\centering
\scalebox{0.70}{
\begin{tabular}{c|l|cccc|cccccccc}  
\toprule
\multicolumn{1}{c}{} & & \textit{gate} & \textit{pseudoL} & \textit{unique} & \textit{image} & BLEU-1 & BLEU-2 & BLEU-3 & BLEU-4 & METEOR & ROUGE-L & CIDEr & SPICE \\
\midrule
\multirow{5}{*}{A} & Ours (full) & \checkmark & \checkmark & \checkmark & \checkmark & \textbf{49.5} & \textbf{27.3} & \textbf{13.1} & 6.3 & 14.0 & 34.5 & \textbf{31.9} & \textbf{8.6} \\
                   & w/o \textit{pseudoL} & \checkmark &  & \checkmark & \checkmark & 0.0 & 0.0 & 0.0 & 0.0 & 0.0 & 0.5 & 0.9 & 0.3 \\
                   & w/o \textit{gate} &  &  & \checkmark & \checkmark & 40.9 & 21.5& 10.1 & 4.8 & 12.7 & 32.1 & 17.6 & 6.0 \\
                   & w/o \textit{unique} & \checkmark & \checkmark &  & \checkmark & 47.2 & 26.2 & 13.0 & \textbf{6.4} & \textbf{14.1} & \textbf{34.9} & 28.3 & 8.5 \\
                   & w/o \textit{image} & \checkmark & \checkmark & \checkmark &  & 43.3 & 23.3 & 10.8 & 5.1 & 13.1 & 31.7 & 25.5 & 7.8 \\
\midrule\midrule
% \midrule
\multirow{5}{*}{B} & Ours (full) & \checkmark & \checkmark & \checkmark & \checkmark & \textbf{50.4} & \textbf{29.5} & \textbf{14.4} & \textbf{7.6} & \textbf{13.5} & \textbf{37.3} & \textbf{31.8} & \textbf{8.4} \\
                   & w/o \textit{pseudoL} & \checkmark &  & \checkmark & \checkmark & 44.5 & 25.4 & 12.2 & 6.2 & 12.4 & 36.7 & 29.2 & 7.5 \\
                   & w/o \textit{gate} &  &  & \checkmark & \checkmark & 44.5 & 24.2 & 12.0 & 6.2 & 11.6 & 34.2 & 19.4 & 5.8 \\
                   & w/o \textit{unique} & \checkmark & \checkmark &  & \checkmark & 47.9 & 27.1 & 13.0 & 6.4 & 12.6 & 36.3 & 26.9 & 7.4 \\
                   & w/o \textit{image} & \checkmark & \checkmark & \checkmark &  & 47.1 & 26.0 & 12.8 & 6.6 & 13.1 & 34.7 & 29.7 & 8.0 \\
\bottomrule
\end{tabular}
}
\caption{Ablation studies on the experimental settings A and B. The scores of Ours (full) are the mean of five runs; those of the other ablated models are the results of a single run.}
\label{tab:ablation}
\end{table*}

\begin{table*}[t]
\centering
\scalebox{0.77}{
\begin{tabular}{lcccccccc}  
\toprule
 & BLEU-1 & BLEU-2 & BLEU-3 & BLEU-4 & METEOR & ROUGE-L & CIDEr & SPICE \\
\midrule
\newcite{feng2019} & 41.0 & 22.5 & 11.2 & 5.6 & 12.4 & 28.7 & 28.6 & 8.1 \\
% \midrule
Ours & 49.5 $\pm$ 0.7 & \textbf{27.3 $\pm$ 1.2} & \textbf{13.1 $\pm$ 0.8} & 6.3 $\pm$ 0.5 & \textbf{14.0 $\pm$ 0.1} & 34.5 $\pm$ 0.3 & 31.9 $\pm$ 1.0 & 8.6 $\pm$ 0.2 \\
% \midrule
Ours + \newcite{feng2019} & \textbf{50.9 $\pm$ 0.1} & \textbf{28.0 $\pm$ 0.1} & \textbf{14.0 $\pm$ 0.1} & \textbf{7.1 $\pm$ 0.0} & \textbf{14.1 $\pm$ 0.0} & \textbf{35.2 $\pm$ 0.1} & \textbf{35.7 $\pm$ 0.1} & \textbf{9.2 $\pm$ 0.0} \\
\bottomrule
\end{tabular}
}
\caption{Results of combining our method with previous methods \cite{feng2019}. Scores of our model and the combined model are the \textit{mean} $\pm$ \textit{standard deviation} of five runs. We marked in bold the scores within the standard deviation of the best scores.}
\label{tab:refinement}
\end{table*}

\subsection{Comparison with the State-of-the-Art Results}

Table~\ref{tab:sota comparison} lists the results of our model compared with the previous state-of-the-art results.
To avoid evaluating cherry-picked scores, we computed the mean and standard deviation of five results obtained with different seeds\footnote{In all the experiments, we specified a seed of 0, 1, 2, 3, 4 for each run.}.
Our method outperforms the previous approaches in terms of all evaluation metrics.
These results confirm the effectiveness of our simple method.

\subsection{Ablation Study}

Table~\ref{tab:ablation} lists the results of our model obtained in the ablation studies.
We tested the ablation of the gating mechanism (\textit{gate}), pseudo-labels on the gating mechanism (\textit{pseudoL}), unique-object decoding (\textit{unique}), and image features (\textit{image}).
The pseudo-labels cannot be implemented without the base gating mechanism.
Thus, the model ``w/o \textit{gate} w/ \textit{pseudoL}'' is not applicable.
The model w/o \textit{image} is the same as Ours (full) except that it only uses the word embeddings of detected object labels, rather than image features.
It encodes detected object labels into word embeddings and then takes their mean\footnote{The number of detected objects was 3.0 in setting A and 4.0 in setting B on average. Thus, taking the mean does not break the detected information significantly.} and replaces the image feature $\bm{v}$ with it.
All models here were trained in the same manner as described in Section~\ref{implement}.

The results show that the pseudo-labels on the gating mechanism contribute a lot to the performance; the score degrades significantly from Ours (full) to w/o \textit{pseudoL} in all metrics.
On the other hand, the base gating mechanism does not function well by itself; not all scores of w/o \textit{gate} are lower than those of w/o \textit{pseudoL}.
These results demonstrate that explicitly removing the word-level spurious alignment contributes the most to the relatively high performance of our model.
Although it is a relatively low contribution compared with the pseudo-labels, unique-object decoding also enhances performance.

The degraded performance of w/o \textit{image} suggests that object labels themselves are insufficient to describe images correctly.
We observed that this model was vulnerable to errors propagated through object detectors.
See Section~\ref{qualitetive analysis} for the examples.

\subsection{Combining with Previous Methods}
\label{combine}

Our method focuses on removing word-level spurious alignment between images and pseudo-captions, whereas the previous methods focus on aligning images and pseudo-captions at the sentence level.
To utilize the strength of each, we combined our method with the previous methods as a model initialization method.

We first trained our model on the setting A and generated captions for the images in training data.
We then paired the captions with the images as their pseudo-captions\footnote{To avoid assigning obviously incorrect pseudo-captions, we omitted the captions that contained fewer than one detected object for the images with more than two detected objects.
For the images with fewer than one detected object, we omitted the captions that contained no detected objects.}.
With the pairs, the caption generator of \newcite{feng2019} was initialized by learning to generate the pseudo-captions from the images.
After the initialization, we trained the previous model using their publicly available code\footnote{\url{https://github.com/fengyang0317/unsupervised_captioning}}.
We used the same hyperparameters as \newcite{feng2019} except for the learning rate of $10^{-5}$ for the generator and $10^{-8}$ for the discriminator.

Table~\ref{tab:refinement} shows the results.
The combined model further improves the performance of our model and \newcite{feng2019}.
In particular, the improvement from \newcite{feng2019} is much larger than that from our model.
These results suggest that removing the word-level spurious alignment is critical for the subsequent sentence-level alignment.

\begin{table}[t]
\centering
\scalebox{0.72}{
\begin{tabular}{llccc}  
\toprule
\multicolumn{1}{c}{} & & Precision & Recall & F1 \\
\midrule
\multirow{3}{*}{Detected} & \newcite{feng2019} & \textbf{56.6} & 57.4 & \textbf{55.4} \\
% \cline{2-5}
                  & Ours & 51.0 & 56.7 & 51.6 \\
% \cline{2-5}
                  & Ours + \newcite{feng2019} & 54.0 & \textbf{61.8} & \textbf{55.4} \\

\midrule
\multirow{3}{*}{Others} & \newcite{feng2019} & 22.3 & 17.0 & 18.8 \\
% \cline{2-5}
                  & Ours & 27.8 & \textbf{21.9} & 23.4 \\
% \cline{2-5}
                    & Ours + \newcite{feng2019} & \textbf{29.9} & \textbf{21.9} & \textbf{24.2} \\
\bottomrule
\end{tabular}
}
\caption{Bag-of-words matching scores with respect to detected object labels and the other words.}
\label{tab:object bleu}
\end{table}

\subsection{Negative Effect of Spurious Alignment}
\label{obj eval}

To further investigate the effect of removing the spurious alignment, we evaluated our model on \textit{noisier words}: words other than the detected object labels.
Our method discourages from aligning them with images because they are likely to be irrelevant to given images, while previous methods force the alignment.
We tested the following bag-of-words matching on the MS COCO test set.

Let $S$ be the bag of words of a caption generated from an image $\bm{I}$ and $T^m$ be the $m$-th reference caption of $\bm{I}$.
Given a set of detected object labels $\mathcal{O}$ of $\bm{I}$, we took the intersections $S_{det} = S \cap \mathcal{O}$ and $S_{other} = S \cap \overline{\mathcal{O}}$ for $S$, as well as for $T^m$.
We define the precision ($P$), recall ($R$) and F1 score ($F$) of $S$ against $T^m$ as follows: $P = \frac{|S \cap T^m|}{|S|}$, $R = \frac{|S \cap T^m|}{|T^m|}$, $F = 2 \cdot \frac{P \cdot R}{P + R}$.
Based on this, we define the precision, recall, and F1 score of $S_{det}$ against $T^m_{det}$ by replacing $S$ with $S_{det}$ and $T^m$ with $T^m_{det}$, and likewise for those of $S_{other}$ against $T^m_{other}$.
We calculated the above scores for each pair of a generated captions and their reference captions and subsequently averaged it across the pairs.
The pairs with empty $T^m_*$ were excluded from the calculation.

Table~\ref{tab:object bleu} shows the results.
Overall, the scores on detected object labels (Detected) are about two times higher than those on the other words (Others), indicating the difficulty of learning the alignment of the latter, noisier words.
Our model performs better in predicting the noisier words, outperforming \newcite{feng2019} in all metrics.
These results indicate that refraining from the alignment works better than forcing it for the noisier words.

On the other hand, our model performs worse in predicting detected object labels.
This is because our method trusts all detected object labels and aligns them with images without any constraints used in previous work.
Combined with the previous method (Ours + \newcite{feng2019}), our model improves the prediction on detected object labels.

\begin{table}[t]
\centering
\scalebox{0.75}{
\begin{tabular}{llcc}  
\toprule
\multicolumn{1}{c}{} & & Word Type & Frequency  \\
\midrule
\multirow{3}{*}{Object} & \newcite{feng2019} & 205 & 20013 \\
                  & Ours & 306 & 15052 \\
                  & Ours + \newcite{feng2019} & 239 & 18226 \\

\midrule
\multirow{3}{*}{Others} & \newcite{feng2019} & 827 & 24865 \\
                  & Ours & 169 & 83693 \\
                    & Ours + \newcite{feng2019} & 121 & 110358 \\
\bottomrule
\end{tabular}
}
\caption{Analysis of generated captions with respect to object labels and the other words. Word Type is the number of unique words, and Frequency is the mean of the frequency of the words in the training text corpus.}
\label{tab:stats}
\end{table}

\begin{figure*}[t]
    \centering
    \includegraphics[width=1.0\textwidth,keepaspectratio]{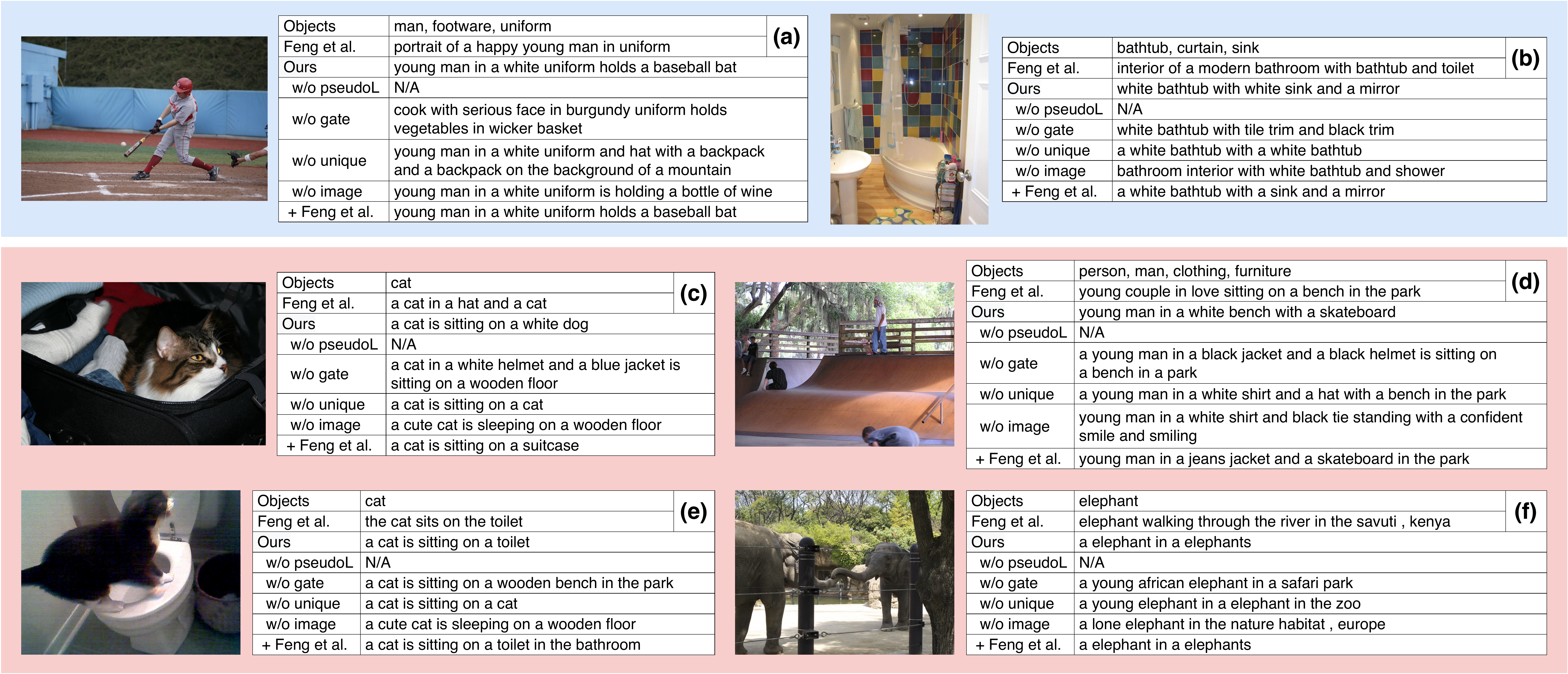}
    \caption{Sample captions for six input images taken from the MS COCO validation set. Our model generated correct captions for the images in the top row and wrong captions for the rest. Best viewed by zooming in.}
    \label{fig:caption}
\end{figure*}

\subsection{Positive Effect of Frequency}

By assigning the pseudo-label $f$, our method encourages to align detected object labels with the image representation $\bm{v}$ and the other words with the contextual representation $\bm{h}$.
Thus, our model is likely to predict the latter words mostly based on the previous output sequences, as language models do.
If this is the case, then the latter words predicted tend to be the frequent words in the training text corpus.

To verify this tendency, we analyzed the frequency of output words in the training text corpus for object labels and the other words\footnote{As we analyzed each unique word across all output captions in the MS COCO test set, we roughly divided the words into object labels and the others, not into detected object labels and the others.}.
Table~\ref{tab:stats} presents the results.
In contrast to object labels, our outputs' vocabulary is about five times smaller than that of \newcite{feng2019}, and the words tend to be highly frequent in the training text corpus.

The results also show that a model performs better if it has the smaller and more frequent vocabulary of the words other than object labels.
This correlation is convincing considering the coverage of frequent words.
For example, a general caption such as ``a man \textit{with} a bike'' can correctly cover various scenes in which a man is riding/sitting on/leaning on/standing near/... a bike.
This positive effect of frequency suggests that firstly aligning the frequent words and gradually extending them can be a promising approach.

\subsection{Qualitative Analysis on Outputs}
\label{qualitetive analysis}

Figure~\ref{fig:caption} shows the captions generated by our model, its ablated models, \newcite{feng2019}, and the combined model trained on the setting A.
Our model generated correct captions for images (a) and (b).
It successfully generated object labels that were not even detected by the object detector: \textit{bat} in (a) and \textit{mirror} in (b).
On the other hand, errors of the object detector directly propagated to the output captions of w/o \textit{image} model: the model generated an incorrect object \textit{a bottle of wine}, owing to the missing object \textit{bat} in (a).

Captions of the other images are negative results of our model.
We observed that our model tended to repeat similar objects: \textit{cat} and \textit{dog} in (c), and \textit{elephant} and \textit{elephants} in (f).
Without unique-object decoding, this tendency got worse: w/o \textit{unique} model repeated \textit{cat} in (c) and (e), and \textit{elephant} in (f).
Ours + \newcite{feng2019} model did not change much of the prediction of our model, as we set the learning rate low (see Section~\ref{combine}).
However, it allowed the partial correction seen in (c): the combined model modified \textit{dog } to \textit{suitcase}.

In our outputs, words other than object labels tended to be frequent words and composed short phrases.
On the contrary, \newcite{feng2019} tended to generate less frequent words (\textit{savuti} and \textit{kenya} in (f)) and longer phrases (\textit{portrait of a happy young} in (a) and \textit{young couple in love} in (d)), which were incorrect predictions in these examples.

\begin{figure}[t]
    \centering
    \includegraphics[width=1.0\columnwidth,keepaspectratio]{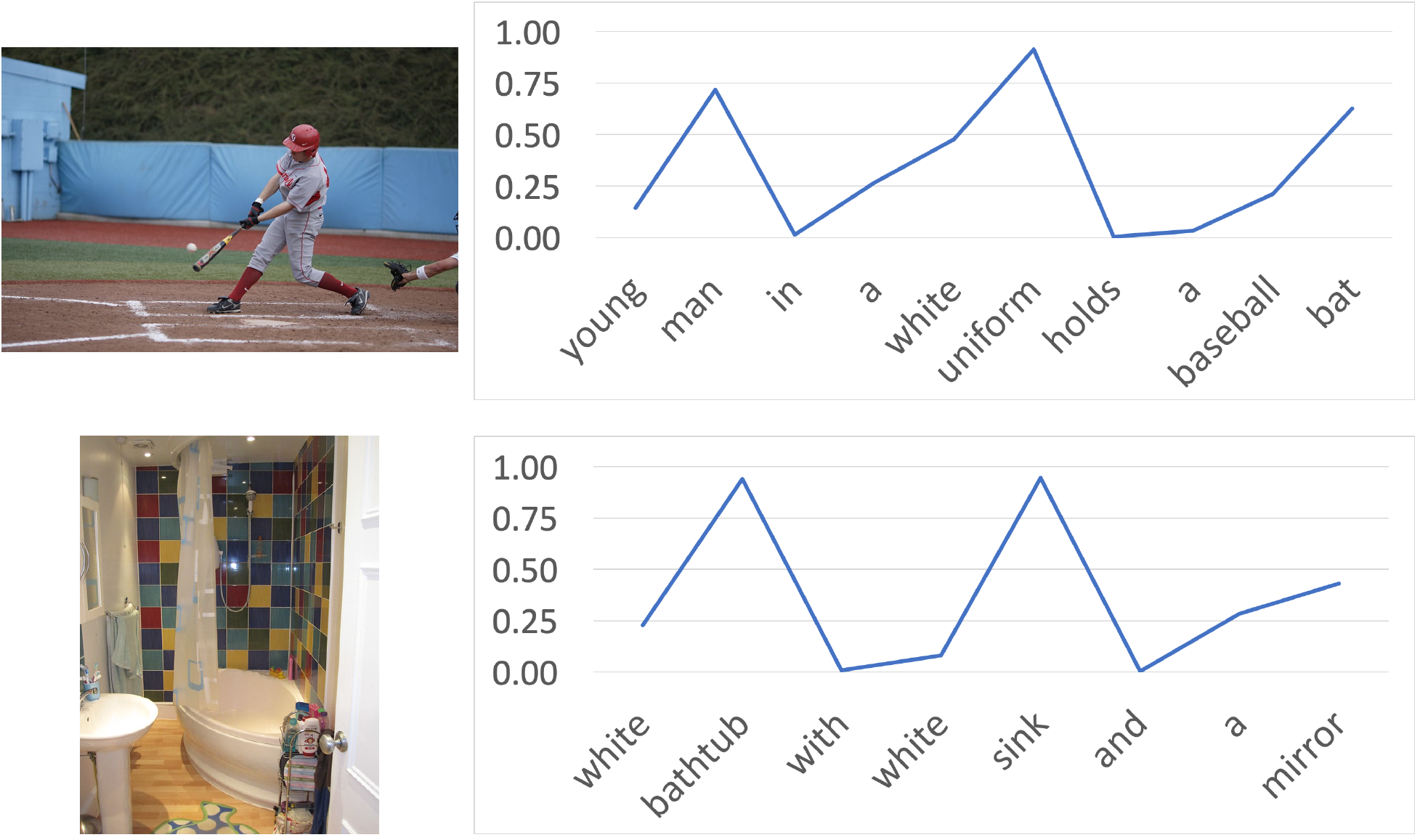}
     \caption{Sample captions with gate values. The plot represents the values of $g_t$ for each predicted word. The value of $g_t$ becomes high when the word is predicted using mainly image representation.}
    \label{fig:gate caption}
\end{figure}

Figure~\ref{fig:gate caption} shows output captions of our model and the gate values for each word.
Overall, the gate values were high for object labels and low for the other words.
Although our model was correct on the words other than object labels in these examples, these words were generated mostly by contextual features, thus heavily relied on contextual frequency.
This heavy reliance on contexts resulted in generating the same word after an object label without considering images: \textit{is sitting on} followed \textit{cat} in both (c) and (e).

\section{Related Work}

There has been considerable research with different settings and approaches to describe scenes that have no image--sentence pairs.
Novel object captioning \cite{hendricks2016deep,venugopalan2017captioning, anderson2018partially, agrawal2019nocaps} attempted describing unseen objects in captions.
They incorporated an image classifier or object detector trained on objects not included in image--sentence pairs.
\newcite{lu2018} tested captioning models on the generation of unseen combinations of objects, and \newcite{nikolaus2019} extended this to the unseen combinations of objects, attributes, and relations.
In both settings, only the combinations were unseen, but each word in the combinations appeared in the training data.
Semi-supervised approaches utilized caption retrieval models to automatically collect the corresponding captions for unannotated images to augment image--sentence pairs \cite{liu2018show, kim2019}.

The above work was evaluated on the scenes where correct descriptions partially overlapped with those in the training image--sentence pairs.
However, there can be scenes with no such overlap due to the limited coverage of the currently available image--sentence pairs.
Taking a step further, unsupervised image captioning \cite{feng2019, laina2019} aims to describe scenes that have no overlap with the image--sentence pairs, without the annotation of the pairs.
To test in that situation, the task does not allow to use any image--sentence pairs.
The only available resources are images and sentences drawn from different sources and object labels detected from the images.

\newcite{feng2019} first trained an encoder--decoder model that takes object labels in a sentence as its input and outputs the original sentence.
After training, this model took the object labels detected from each image and outputted a sentence to pair with the image as its pseudo-caption.
These pairs were then used to initialize a caption generator for the subsequent image--sentence alignment: bi-directional (image-to-sentence and sentence-to-image) feature reconstruction and GAN training \cite{goodfellow2014} to ensure fluency in generated captions.
In the work of \newcite{laina2019}, pseudo-captions were sentences that contained object labels detected from a given image.
They employed metric learning and GAN training to minimize the difference between images and pseudo-captions in their latent space, as well as to maximize the difference between images and sentences wherein no detected object label was included.
In concurrent work, \newcite{cao2020} introduced an additional attention network into the model of \newcite{feng2019}.
They pre-trained the attention network on extra image resources annotated with objects and relations of the objects so that it could consider the interactions between the objects.

Our approach is different from them in that it focuses on removing the mismatched words of pseudo-captions to take reliable supervision only, rather than forcing the use of the entire pseudo-captions for image--sentence alignment.
Although the previous work additionally ensured to align detected object labels to images, they did not prevent the spurious alignment between images and words.

\newcite{guo2020} is a concurrent work that proposed a memory network to generate natural sentences from detected object labels.
They focused on filling the gap between a set of discrete words and natural sentences.
Our work and theirs are the same in that the focus is not on image--sentence alignment at the sentence level; the difference is that we focus on investigating the effect of removing word-level spurious alignment.
We designed our method simply and explicitly for the objective and provided in-depth analyses on the effect.

As an eased setting of unsupervised image captioning, unpaired image captioning has also been explored \cite{feng2019,laina2019,gu2019,liu2019}.
The major difference from unsupervised image captioning is that images and sentences are drawn from image--sentence pairs, rather than from different sources.
That is, every image has completely matched captions in pseudo-captions, which is not the case in unsupervised image captioning.
As correct captions exist for each image, previous approaches focused on matching images and sentences at the sentence level.
Contrary to these approaches, we focus on employing unsupervised image captioning and devising a method to remove word-level spurious alignment in the much noisier pseudo-captions.

Another variation of unpaired image captioning is the generation of captions in one language that has no image--sentence pairs, using paired images and captions in another language \cite{gu2018,song2019}. 
However, this line of research is beyond the scope of our work, as it requires image--sentence pairs to be at least in one language.

Our gating mechanism borrowed the idea of adaptive attention \cite{lu2017,lu2018}.
Adaptive attention serves to control when generating words from image representations.
Although these methods assume that the control is automatically learned from image--sentence pairs, this is not the case in an unsupervised setting.
Our method is different from theirs in that we add heuristic pseudo-labels to train the gate when using image representations.

\section{Conclusion}

We investigated the importance of removing word-level spurious alignment between images and pseudo-captions in the task of unsupervised image captioning.
For this purpose, we introduced a simple gating mechanism trained to align image features with only the most reliable words in pseudo-captions.
The experimental results showed that our proposed method outperformed the previous methods without the sentence-level learning objectives used in the previous methods.
Moreover, our method improved the performance further by combining with the previous methods.
These results confirmed the importance of careful alignment in word-level details.

\bibliography{anthology,eacl2021}
\bibliographystyle{acl_natbib}

\end{document}